\documentclass{article}
\usepackage{PRIMEarxiv}
\usepackage[utf8]{inputenc} 
\usepackage[T1]{fontenc}    
\usepackage{palatino,eulervm}
\usepackage{hyperref}       
\usepackage{url}            
\usepackage{booktabs}       
\usepackage{amsfonts}       
\usepackage{nicefrac}       
\usepackage{microtype}      
\usepackage{lipsum}
\usepackage{fancyhdr}       
\usepackage{graphicx}       
\graphicspath{{media/}}     

\title{Kalman Filters and Homography: Utilizing the Matrix $A$}

\author{
  Burak Bayramlı \\
  İstanbul, Turkey\\
  \texttt{burakbayramli.github.io} 
}

\begin{document}
\maketitle

\begin{abstract}
Many problems in Computer Vision can be reduced to either working around a known
transform, or given a model for the transform computing the inverse problem of
the transform itself. We will look at two ways of working with the matrix $A$
and see how transforms are at the root of image processing and vision problems.
\end{abstract}

\keywords{Homography \and Kalman Filters \and Computer Vision}

\section{Introduction}

As stated by a great teacher of linear algebra identifying the $A$ is crucially
important in many Applied Mathematics problems \cite{strang} because linear
transformation is a key abstraction, we know everything when we know what
happens to a basis. Many applications, from stress analysis to network analysis
involves the construction of an $A$, i.e. a transformation that takes certain
inputs and maps them to certain outputs.

\section{Direct Linear Transform}

In computer vision a transform between two 2D pictures that can involve a
translation, scaling and a rotation is named a homography, and can be stated
simply as;

$$ x' = H x$$

where $x,x'$ 2D pixel coordinates. In expanded form we have \cite{solem},

$$ 
\left[\begin{array}{r} x' \\ y' \\ w' \end{array}\right]
\left[\begin{array}{rrr}
h_1 & h_2 & h_3 \\
h_4 & h_5 & h_6 \\
h_7 & h_8 & h_9 
\end{array}\right]
\left[\begin{array}{r} x \\ y \\ w \end{array}\right]
$$

Homogeneous coordinates are used. The use of letter $H$ is the custom, but the
transformation above in this problem is our $A$.

A special case of transformation, so-called affine transformation is shown
below,

$$ 
\left[\begin{array}{r} x' \\ y' \\ 1 \end{array}\right]
\left[\begin{array}{rrr}
a_1 & a_2 & t_x \\ a_3 & a_4 & t_y \\ 0 & 0 & 1
\end{array}\right]
\left[\begin{array}{r} x \\ y \\ 1 \end{array}\right],
\qquad
x' = \left[\begin{array}{rr} A & t \\ 0 & 1 \end{array}\right] x
$$

This transformation preserves $w = 1$ and for example the translation vector
inside it is at $[t_x, t_y]$.

A similarity transformation is,

$$ 
\left[\begin{array}{r} x' \\ y' \\ 1 \end{array}\right]
\left[\begin{array}{rrr}
s\cos(\theta) & -s\sin(\theta) & t_x \\ 
s\sin(\theta) & s\cos(\theta) & t_y \\ 
0 & 0 & 1
\end{array}\right]
\left[\begin{array}{r} x \\ y \\ 1 \end{array}\right],
\qquad
x' = \left[\begin{array}{rr}
sR & t \\ 0 & 1
\end{array}\right] x
$$

Similarity transformations can include scale changes as well, rotation is
captured with the cell elements containing $\sin$, $\cos$.

Now we arrive at the reverse problem; how, based on a few matching point between
two images, one original the other transformed, do we find the homography, the
transformation that led to the resulting matrix?

We assume the matching points are in the form of $x_i,y_i$ and $x_i',y_i'$.  If
we expand $x' - Hx = 0$ for every such matches we obtain,

$$ 
\left[\begin{array}{rrrrrrrrr}
-x_1 & -y_1 & -1 & 0 & 0 & 0 & x_1x_1' & y_1x_1' & x_1' \\
0 & 0 & 0 & -x_1 & -y_1 & -1 & x_1y_1' & y_1y_1' & y_1' \\
-x_2 & -y_2 & -1 & 0 & 0 & 0 & x_2x_2' & y_2x_2' & x_2' \\
0 & 0 & 0 & -x_2 & -y_2 & -1 & x_2y_2' & y_2y_2' & y_2' \\
 &  \vdots &  &  \vdots &  & \vdots &  &  \vdots & 
\end{array}\right]
\left[\begin{array}{r}
h_1 \\ h_2 \\ h_3 \\ h_4 \\ h_5 \\ h_6 \\ h_7 \\ h_8 \\ h_9 
\end{array}\right] = 0
$$

More matches as data points would allow the expansion of the matrix vertically.
The reason for representing $x' - Hx = 0$ instead of $x'=Hx$ is we can see the
former as an optimization problem, even if we can't solve for equality to zero,
we can attempt to {\em approach} zero, by using a singular value decomposition
method.

A fantastic and simple application of DLT can be demonstrated even without even
known matches per se between two images, if we want certain section of an image
to be extracted and scaled, rotated into a full-blown straight image, we could
simply define four corners of the selection area to be the outermost corners of
an imaginary target output. Let's consider the Sudoku image below,

\begin{figure}[h]
  \centering
  \includegraphics[width=10em]{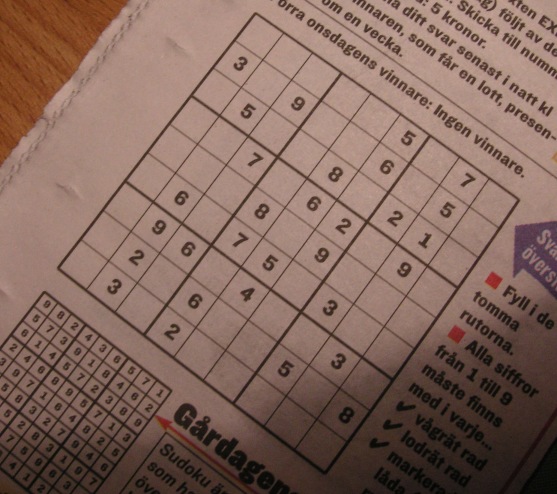}
\end{figure}

If we want to extract the sudoku section of that image,

\begin{verbatim}
from scipy import ndimage
from PIL import Image

im = np.array(Image.open('sudoku81.JPG').convert('L'))
corners = [[257.4166, 14.9375], 
           [510.8489, 197.6145], 
           [59.30208, 269.65625], 
           [325.598958, 469.05729]]
corners = np.array(corners)
plt.plot(corners[:,0], corners[:,1], 'rd')
plt.imshow(im, cmap=plt.cm.Greys_r)
\end{verbatim}

Here picked corners of the image area are shown in red. Using the SVD
calculation shown in the Appendix, we can obtain an $H$. Now we can warp the
selection using the newly found $H$ and retrieve the full-blown extracted image.

\begin{verbatim}
def warpfcn(x):
    x = np.array([x[0],x[1],1])
    xt = np.dot(H,x)
    xt = xt/xt[2]
    return xt[0],xt[1]
im_g = ndimage.geometric_transform(im,warpfcn,(300,300))
\end{verbatim}

\begin{figure}[h]
  \centering
  \includegraphics[width=22em]{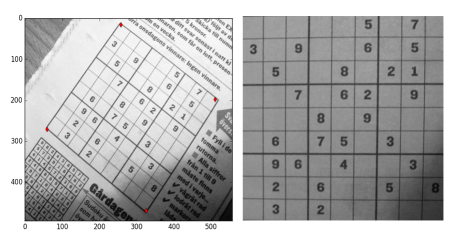}
\end{figure}

\section{Kalman Filters}

Another domain where we ask ``where is the $A$?'' is filtering. But to be exact,
in this case we are asking where is the $A$ and $H$? Kalman filters use two
transformations in succession to compute a final model. The Kalman Filter model
is,

\begin{eqnarray*}
x_{t+1} = A x_t + Q\\
y_t = Hx_t + R
\end{eqnarray*}

where $Q$ and $R$ are multivariate noise. Kalman filters are widely used for
object tracking through noisy measurements. The $x_t$ could be the location of
an object plus noise, $y_t$ could be measurement through visual, radar, sonar
methods and is on a different domain than $x_t$. One obvious example is visual
tracking of an object that moves in 3D space; we see its projection onto a 2D
plane through a certain transformation (image projection) while successive
$x_t$'s are transformed through a displacement transition due to the known
movement mechanics of the object.

In filtering case both $A,H$ can be assumed as known, the aim is usually trying
to ``reverse the arrow'', meaning for given $y_t$ deducing a hidden $x_t$. In
this sense Kalman filters are examples of recursive filters and $x_t$ and $y_t$
can be assumed to be distributed as Gaussians due to the added noise at both
steps.

We can implement a visual tracking using the approaches outlined above. We move
a chessboard plane on a flat surface (table) on constant speed toward our camera
while tracking a single reference point on this image where $x_t$ is the 3D
location of our chessboard plane. The transition equation of the Kalman filters
only needed to account for constant velocity, along one axis. Matrix $A$ would
look like,

\begin{eqnarray*}
A= \left[
\begin{array}{cccc}
1 & 0 & 0 & 0 \\
0 & 1 & 0 & 0 \\
0 & 0 & 1 & d \\
0 & 0 & 0 & 1
\end{array}
\right]
\end{eqnarray*}

Note that matrix A is 4x4, not 3x3. We used homogenous coordinates to represent
3D location, which made $x_t$ a 4x1 vector, therefore the transition equation
captured in $A$ had to be 4x4 so we could multiply it with $x_t$.

\begin{figure}[h]
  \centering
  \includegraphics[width=15em]{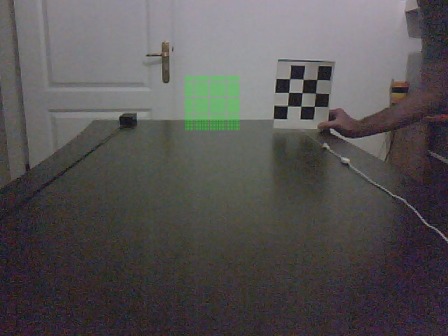}
\end{figure}

$A$ is the identity matrix $I$ with one difference; its (3,4)'th item is a
constant, $d$, that is our displacement. To verify if this $A$ can be used for
calculating displacement, one could perform a sample multiplication using $x_t =
[\begin{array}{cccc}a1 & a2 & a3 & a4 \end{array}]$ and see $Ax_t$, or
$x_{t+1}$, becomes $[\begin{array}{cccc}a1 & a2 & a3+d & a4\end{array}]$. We can
see displacement d is added to the z-coordinate, depth. For our testing we
picked $d = -0.5$, meaning for each frame chessboard plane moved 0.5 cm toward
the camera. Each transition for $x_t$ adds to uncertainty, and that is reflected
in $Q$.

Observation equation $y_t$ handles the calculation where a 3D coordinate is
projected onto 2D (pixel measurements) on screen with added noise. For standard
pinhole camera model, camera matrix $P$ (the $H$ of the KF) is responsible for
2D projection. $P$ is unique for each camera, and a process called camera
calibration can determine camera matrix $P$, and various methods for calibration
exist in literature. OpenCV library contained a function for calibration which
we tested, but we were not happy with the results. At the end, we calibrated the
camera manually; by simply deciding on a 3D real world location (middle point at
the end of our flat surface / table) and tested various $P$ values while at the
same time drawing an imaginery chessboard image on screen using this projection
matrix $P$. This was repeated until the imaginary board was located at the
desired location.

\begin{verbatim}
import cv2

dim = 3
if __name__ == "__main__":    
    fin = sys.argv[1]
    cap = cv2.VideoCapture(fin)
    N = int(cap.get(cv2.CAP_PROP_FRAME_COUNT))
    fps = int(cap.get(cv2.CAP_PROP_FPS))
    kalman = Kalman(util.K, mu_init=array([1., 1., 165., 0.5]))
    
    for i in range(N):
        ret, frame = cap.read()
        h,w = frame.shape[:2]
        gray = cv2.cvtColor(frame, cv2.COLOR_BGR2GRAY)
        status, corners = cv2.findChessboardCorners( gray, (dim,dim))
        is_x = []; is_y = []
        if status: 
            cv2.drawChessboardCorners( gray, (dim,dim), corners, status)
            for p in corners:
                is_x.append(p[0][0])
                is_y.append(p[0][1])

        if len(is_x) > 0 : 
            kalman.update(array([is_x[5], h-is_y[5], 1.]))
            util.proj_board(gray, 
                            kalman.mu_hat[0], 
                            kalman.mu_hat[1], 
                            kalman.mu_hat[2])

        cv2.imshow('frame',gray)
        cv2.waitKey(20)    
\end{verbatim}

The tracking is computed with the code above, standard KF tracking. We used a
chessboard image so that \verb!cvFindChessboardCorners! and
\verb!cvDrawChessboardCorners! OpenCV functions could be used. Using these two
methods, a chessboard image can be detected and marked (on screen, real-time)
with great accuracy and speed. The chessboard image had 9 squares on it, giving
us 9 points of which, we only used the 5th, middle point. At each frame
\verb!cvFindChessboardCorners! detected the 2D pixel locations, and we passed
these values over to the Kalman filter that recursively updated its hidden
state. In each case initial condition for the filter is the middle end point,
somewhat away from both starting locations, therefore the uncertainty $Q$ of the
filter at time = 0 had to be large. Hence we used $Q = I \cdot 150$ cm for the
Kalman filter.

\begin{figure}[h]
  \centering
  \includegraphics[width=30em]{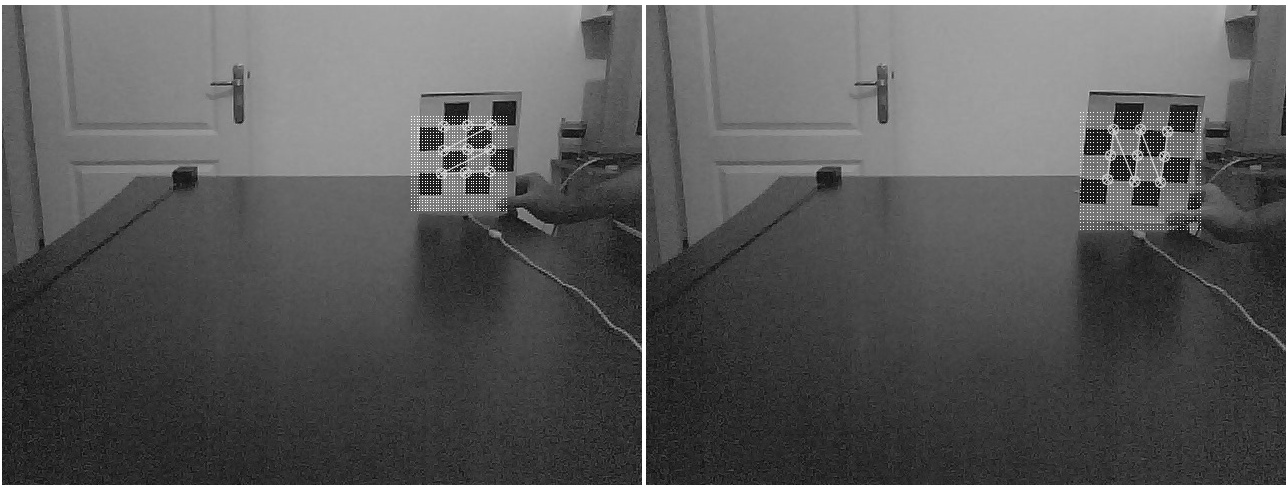}
\end{figure}

\section{Appendix: Code}

\subsection{Homography, SVD}

\begin{verbatim}
import scipy, numpy.linalg as lin
from scipy import ndimage

def H_from_points(fp,tp):
    if fp.shape != tp.shape:
        raise RuntimeError('number of points do not match')
        
    m = np.mean(fp[:2], axis=1)
    maxstd = np.max(np.std(fp[:2], axis=1)) + 1e-9
    C1 = np.diag([1/maxstd, 1/maxstd, 1]) 
    C1[0][2] = -m[0]/maxstd
    C1[1][2] = -m[1]/maxstd
    fp = np.dot(C1,fp)
    
    m = np.mean(tp[:2], axis=1)
    maxstd = np.max(np.std(tp[:2], axis=1)) + 1e-9
    C2 = np.diag([1/maxstd, 1/maxstd, 1])
    C2[0][2] = -m[0]/maxstd
    C2[1][2] = -m[1]/maxstd
    tp = np.dot(C2,tp)
    
    nbr_correspondences = fp.shape[1]
    A = np.zeros((2*nbr_correspondences,9))
    for i in range(nbr_correspondences):        
        A[2*i] = [-fp[0][i],-fp[1][i],-1,0,0,0,
                    tp[0][i]*fp[0][i],tp[0][i]*fp[1][i],tp[0][i]]
        A[2*i+1] = [0,0,0,-fp[0][i],-fp[1][i],-1,
                    tp[1][i]*fp[0][i],tp[1][i]*fp[1][i],tp[1][i]]
    
    U,S,V = lin.svd(A)
    H = V[8].reshape((3,3))    
    
    H = np.dot(lin.inv(C2),np.dot(H,C1))
    
    # normalize and return
    return H / H[2,2]

fp = [ [p[1],p[0],1] for p in corners]
fp = np.array(fp).T
tp = np.array([[0,0,1],[0,300,1],[300,0,1],[300,300,1]]).T
H = H_from_points(tp,fp)
\end{verbatim}

\subsection{Kalman Filter}

\begin{verbatim}
from numpy import *

class Kalman:
    # T is the translation matrix
    # K is the camera matrix calculated by calibration
    def __init__(self, K, mu_init):
        self.ndim = 3
        self.Sigma_x = eye(self.ndim+1)*150
        self.Phi = eye(4)
        self.Phi[2,3] = -0.5
        self.H = append(K, [[0], [0], [0]], axis=1)
        self.mu_hat = mu_init
        self.cov = eye(self.ndim+1)
        self.R = eye(self.ndim)*1.5
        
    def normalize_2d(self, x): 
        return array([x[0]/x[2], x[1]/x[2], 1.0])
    
    def update(self, obs):

        # Make prediction
        self.mu_hat_est = dot(self.Phi,self.mu_hat) 
        prod = dot(self.Phi, dot(self.cov, transpose(self.Phi)))
        self.cov_est = prod + self.Sigma_x
                
        # Update estimate
        prod = self.normalize_2d(dot(self.H,self.mu_hat_est))
        self.error_mu = obs - prod
        
        prod = dot(self.cov,transpose(self.H))
        prod = dot(self.H,prod)
        self.error_cov = prod + self.R
        prod = dot(self.cov_est,transpose(self.H))
        self.K = dot(prod,linalg.inv(self.error_cov))
        self.mu_hat = self.mu_hat_est + dot(self.K,self.error_mu)
        
        prod = dot(self.K,self.H)
        left = eye(self.ndim+1) 
        diff = left - prod
        self.cov = dot(diff, self.cov_est)
\end{verbatim}

\bibliographystyle{unsrt}

\end{document}